\pdfoutput=1
\documentclass{article}

\usepackage[preprint]{neurips_2024}

\usepackage[utf8]{inputenc}
\usepackage[T1]{fontenc}
\usepackage{xcolor}
\usepackage[colorlinks=true,linkcolor=blue!55!black,citecolor=blue!55!black,
            urlcolor=blue!55!black]{hyperref}
\usepackage{booktabs}
\usepackage{float}
\usepackage[most]{tcolorbox}
\newtcolorbox{promptbox}[1]{enhanced, breakable, colback=black!4, colframe=black!12, boxrule=0.3pt,
  arc=2pt, left=6pt, right=6pt, top=4pt, bottom=4pt, before skip=6pt, after skip=8pt,
  fontupper=\small, title=#1, fonttitle=\small\bfseries, coltitle=black, colbacktitle=black!9,
  titlerule=0pt, toptitle=2pt, bottomtitle=2pt}
\usepackage{multirow}
\usepackage{graphicx}
\usepackage{amsmath}
\usepackage{amssymb}
\usepackage{enumitem}

\makeatletter\renewcommand{\@noticestring}{Preprint.}\makeatother

\title{An LLM Is All You Need:\\
Data-Efficient Adaptation to Full-Duplex, Word-Level Spoken Dialogue}
\author{Tairan Chen, Richard Yucheng He, Yihang Liu, Baodong Cao, Chen Xu\\[6pt]
        \textbf{AveraLabs}\\[4pt]
        {\small\normalfont\texttt{\{terrence.chen, richard, robbie.liu, baodong.cao, chen.xu\}@averalabs.com}}}

\begin{document}
\maketitle

\begin{abstract}
Full-duplex spoken dialogue, in which a system listens while it speaks and can backchannel,
yield the floor, and be interrupted, is usually obtained by training on hundreds of thousands to
millions of hours of speech. This is costly, rewrites the language model that supplies the
system's knowledge, and still leaves turn-taking either implicit or decided coarsely rather than
word by word. We show that a frozen, general-purpose multimodal LLM can instead be adapted into
a word-level full-duplex system with little data. Our system is a thinker--talker cascade on an
80\,ms frame clock. The thinker keeps the LLM frozen and trains only lightweight adapters and a
few new token embeddings (97M parameters); at every frame it emits one token of agent text, with
pauses, yields, and onsets represented as tokens of the same vocabulary and kept legal by a
finite-state machine. A four-stage curriculum, each stage anchored to the model's own behavior
under the previous format, performs the conversion in about 30 GPU-hours. The talker reads only
the thinker's discrete tokens, so the two halves are trained independently. On Full-Duplex-Bench
and InteractionBench, the system is competitive with open-source full-duplex systems built
through large-scale speech training: among them, it is the least likely to take the floor when
the user merely pauses, backchannels most often, and leads in spoken-task accuracy, groundedness,
and coherence over long conversations.
\end{abstract}

\begin{figure}[t]
\centering
\includegraphics[width=\linewidth]{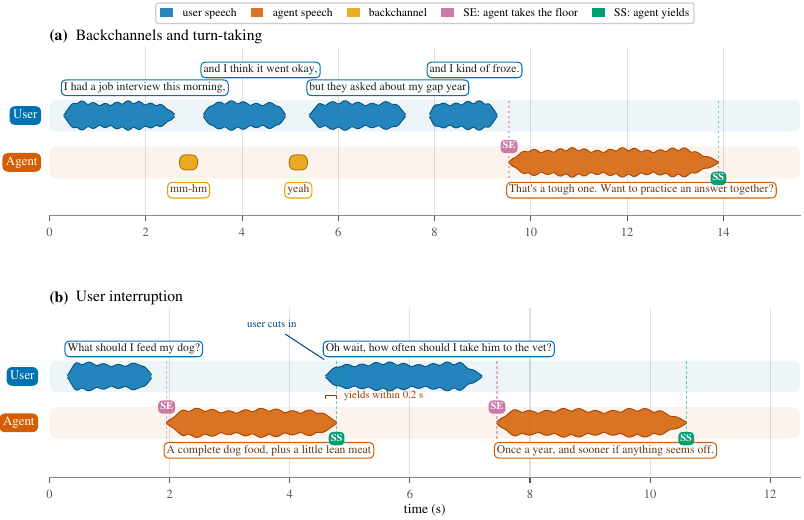}
\caption{Illustration of the two behaviors on a shared timeline (user above, agent below;
schematic: timings and text are illustrative). (a) A free-running conversation: the agent
backchannels (``mm-hm'', ``yeah'', in the backchannel color) while the user is still talking and
takes the turn when the user finishes, with no external turn detector. (b) A user interruption:
the agent yields the floor when the user cuts in (silence-start token, \texttt{SS}) and takes it
back to answer the new request (silence-end token, \texttt{SE}). \texttt{SE}/\texttt{SS}
markers are drawn at the start and end of each agent turn.}
\label{fig:teaser}
\end{figure}

\section{Introduction}
\label{sec:intro}

Spoken dialogue systems built around a turn-based text LLM take turns by construction: the
system waits for the user to finish, transcribes, generates a reply, and speaks it, one role
active at a time. Real conversation is not turn-based. Speakers overlap, interrupt, and
backchannel while the other side is still talking. A full-duplex system listens and speaks on
the same continuous timeline, so interruption (barge-in) and brief acknowledgment (backchannel)
are ordinary events in a signal the model produces and consumes continuously, not special cases
for an external turn detector (Figure~\ref{fig:teaser}). Building one that is both capable and
cheap to obtain remains difficult, for three reasons.

\paragraph{The cost of learning full duplex.} Current systems acquire full-duplex behavior
through very large speech training. Moshi \citep{defossez2024moshi} starts from a 7B-parameter
text LLM pretrained on 2.1T tokens, then pretrains on 7 million hours of transcribed audio before
post-training on diarized multi-speaker audio, the 2{,}000-hour Fisher telephone corpus, and more
than 20{,}000 hours of synthesized instruction dialogue. SyncLLM \citep{veluri2024syncllm}, which
starts from Llama3-8B, trains on 212{,}000 hours of synthetic spoken dialogue. Training at this
scale is expensive, and it rewrites the language model that supplies the system's knowledge;
the authors of Freeze-Omni \citep{wang2024freezeomni} argue that fine-tuning an LLM on limited
speech data risks catastrophic forgetting, and keep it frozen for that reason.

\paragraph{Where turn-taking is decided.} End-to-end models such as dGSLM
\citep{nguyen2022dgslm} and Moshi leave turn-taking implicit in the audio streams they generate,
with no explicit decision that can be inspected or constrained. Systems that make it explicit
decide it coarsely: \citet{zhang2024duplex} process queries and responses in time slices, and
\citet{wang2024fullduplex} let an LLM drive a two-state machine on top of streaming speech
recognition and synthesis, so the model sees the user only as recognized text.
Freeze-Omni shows that a frozen LLM can be given speech input and output with little data, and
adds duplex behavior through an additional multi-task objective. To our knowledge, no prior
system combines a frozen, lightly adapted LLM with turn-taking decided word by word at a fixed
frame rate, including overlap and interruption.
Deciding turn-taking at this granularity raises a problem of its own. A model that samples at
every frame, as frame-level full-duplex models such as Moshi do, decides with each draw whether
to speak, what to say, and whether to pause, and nothing in plain sampling keeps these decisions
consistent: a single ill-placed token, such as a word while the agent is listening or a gap in
the middle of its own turn, yields a sequence unlike any seen in training, from which the model
must then continue.

\paragraph{Coupling between thinking and speaking.} The thinker--talker design separates
deciding what to say from producing speech, but existing instances couple the two halves through
continuous representations. In Qwen2.5-Omni the talker reads the thinker's hidden representations
\citep{xu2025qwen25omni}; Qwen3-Omni stops passing the thinker's text representations but still
feeds the talker its continuous multimodal features and trains and runs the two as one model
\citep{xu2025qwen3omni}. The two halves are therefore trained together.

\begin{figure}[t]
\centering
\includegraphics[width=\linewidth]{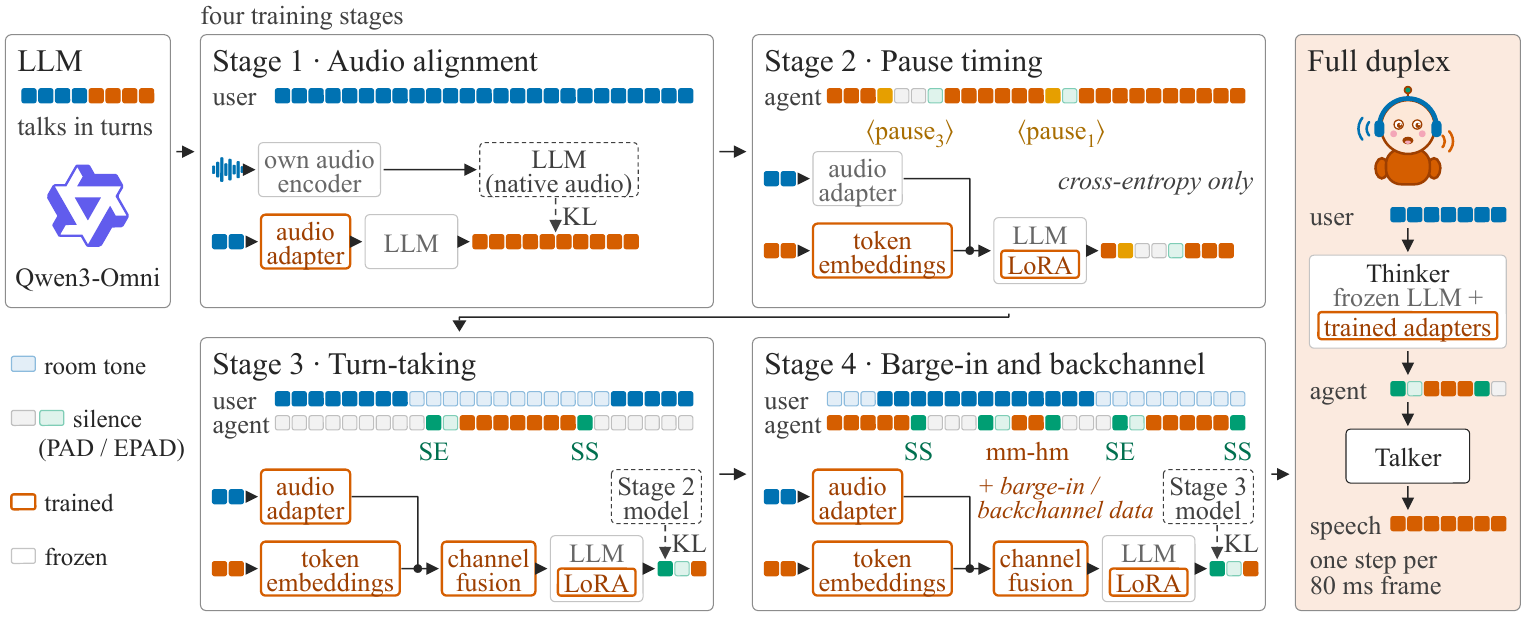}
\caption{Overview. A frozen, general-purpose multimodal LLM (Qwen3-Omni), which only talks in
turns, becomes the thinker of a full-duplex system through four training stages. Each stage
card shows what the model reads and emits at that stage as one cell per frame (user codec
tokens, agent words, pause, \texttt{SE}/\texttt{SS}, and \texttt{PAD} cells), which modules it
trains, and which model anchors its KL term; the frozen core is never changed. The resulting
thinker emits agent text and timing tokens frame by frame, and the talker turns them into
speech, so that the agent listens and speaks at the same time.}
\label{fig:overview}
\end{figure}

\paragraph{Our approach.} This paper addresses these problems by adapting, rather than
retraining, a model that already understands speech and text (Figure~\ref{fig:overview}). We start from Qwen3-Omni
\citep{xu2025qwen3omni}, keep its weights frozen, and convert it into the thinking half of a
full-duplex system by training only low-rank adapters \citep{hu2021lora}, seventeen new token
embeddings, an audio adapter that reads neural-codec tokens, and a channel-fusion layer. On a
12.5\,Hz frame clock, the thinker reads the user's speech and emits one token of the agent's text
per 80\,ms frame; pauses, yields, and onsets are tokens of the same vocabulary, and an explicit
finite-state machine masks illegal tokens before sampling (Section~\ref{sec:thinkerseq}). A
four-stage curriculum (Section~\ref{sec:curriculum}) converts the model, each stage closing one
gap: reading codec audio, timing pauses, taking turns on a two-channel timeline, and handling
interruption and backchannel. No training run uses more than a few tens of thousands of
dialogues, and the whole curriculum takes about 30 GPU-hours on one accelerator. Two design choices aim to keep the stages this
small.
Every new input path is initialized so that the frozen LLM first sees a representation it already
understands (Section~\ref{sec:channelinit}), and each change of format is anchored to the model's
own behavior under the previous one, so that a stage teaches when to speak while holding what
the model says. The talker (Section~\ref{sec:talker}) receives only the thinker's discrete tokens,
so it is trained on speech paired with its word-level transcript, with no thinker in the loop
(Figure~\ref{fig:dataflow}). The whole system is designed around a shared real-time budget: every
80\,ms of audio must be processed within 80\,ms (Section~\ref{sec:framebudget}).

\paragraph{Contributions.} Each contribution answers one of the problems above.
\begin{enumerate}[leftmargin=*,itemsep=1pt,topsep=2pt]
\item A recipe that converts a frozen, general-purpose multimodal LLM into a word-level
full-duplex dialogue model through four training stages totaling about 30 GPU-hours, instead of
large-scale speech training (Section~\ref{sec:training}).
\item A finite-state treatment of turn-taking that keeps per-frame sampling within legal moves:
pause, yield, and onset are tokens of the model's own vocabulary, predicted at every 80\,ms
frame, and a state machine restricts each draw to the tokens that are legal in the current state
(Section~\ref{sec:thinkerseq}).
\item A thinker--talker interface made only of discrete tokens, so that the two halves are
trained separately and in either order (Section~\ref{sec:overview}).
\end{enumerate}
We evaluate on two external benchmarks of real-time behavior, Full-Duplex-Bench
\citep{lin2025fdb,lin2025fdb15} and InteractionBench \citep{averalabs2026interactionbench}
(Section~\ref{sec:setup}). Section~\ref{sec:related} reviews related work,
Sections~\ref{sec:model} and~\ref{sec:training} describe the model and its training,
Sections~\ref{sec:setup} and~\ref{sec:results} the experimental setup and results,
Section~\ref{sec:analysis} analyzes robustness to a noise floor and the talker's intelligibility,
and Sections~\ref{sec:limitations} and~\ref{sec:conclusion} close.

\section{Related Work}
\label{sec:related}

\paragraph{Full-duplex spoken dialogue models.} \citet{nguyen2022dgslm} model two-channel
conversation with dGSLM, a dual-tower transformer with cross-attention over two streams of
discrete speech units, trained on 2{,}000 hours of unlabeled telephone audio with no text
supervision; turn-taking emerges implicitly from the audio rather than being represented by any
explicit signal. Moshi \citep{defossez2024moshi} makes the same parallel-stream idea
text-aware: a single model predicts, at a fixed frame rate, its own text tokens together with
codec tokens for both speakers, again with no explicit turn boundary, and reaches that behavior
through the multi-stage audio pretraining described in Section~\ref{sec:intro}. PersonaPlex
\citep{roy2026personaplex} fine-tunes Moshi to add control of the agent's role by a text prompt
and of its voice by a short reference recording. Other work
adds full-duplex behavior to existing models. LSLM \citep{ma2024lslm} fuses a streaming
listening channel into a decoder-only text-to-speech model and studies where the fusion should
happen. SyncLLM \citep{veluri2024syncllm} adds time information to Llama3-8B so that it runs
synchronously with a real-world clock, training on 212{,}000 hours of synthetic spoken dialogue
and 2{,}000 hours of real dialogue. OmniFlatten \citep{zhang2024omniflatten} progressively
post-trains a text LLM through modality alignment, half-duplex dialogue, and full-duplex
dialogue, standardizing every stage's data with a flattening operation so that the backbone's
architecture is left unchanged. \citet{zhang2024duplex} divide queries and responses into time
slices processed by time-division multiplexing, so that a text LLM can take in new input while it
is still generating. \citet{wang2024fullduplex} keep a cascade of streaming ASR, an LLM, and
streaming TTS, and let the LLM emit control tokens to a two-state machine that decides when to
start responding, wait, or interrupt. PSLM \citep{mitsui2024pslm} is not full duplex, but lowers
response latency by generating text and speech tokens in parallel from one LLM.

\paragraph{Adapting an LLM with little data.} Our thesis is closest to Freeze-Omni
\citep{wang2024freezeomni}, which connects speech input and output to a text LLM whose
parameters stay frozen, trains in three stages on ASR and TTS data plus 60{,}000 multi-round
text question-answer pairs, and adds duplex behavior through an additional multi-task training
objective; its authors argue that keeping the LLM frozen avoids the catastrophic forgetting
that limited data would otherwise cause. We share the frozen-LLM premise with Freeze-Omni and
the staged progression with OmniFlatten. We differ in where turn-taking lives: our thinker emits
word-level agent text on an 80\,ms frame clock, and pauses, yields, and onsets are tokens of
that same stream, learned by the same next-token objective as the words. The adapters
themselves are standard low-rank adapters \citep{hu2021lora}.

\paragraph{Thinker--talker models and neural audio codecs.} The thinker--talker split comes from
Qwen2.5-Omni \citep{xu2025qwen25omni}, where the thinker is an LLM generating text and the
talker is a dual-track autoregressive model that reads the thinker's hidden representations to
produce speech tokens. Qwen3-Omni \citep{xu2025qwen3omni}, whose thinker we adapt, makes both
halves mixture-of-experts models and stops feeding the talker the thinker's text
representations, arguing that for text, discrete tokens and embeddings carry the same
information; its talker still receives the thinker's continuous multimodal features, and the
two are trained and run as one model. Our talker receives only discrete text tokens, so the two
halves are trained separately. For audio tokens we reuse
Mimi \citep{defossez2024moshi}, which descends from residual-vector-quantized codecs:
SoundStream \citep{zeghidour2021soundstream} introduced the encoder--quantizer--decoder design
and EnCodec \citep{defossez2022encodec} extended it to higher fidelity. Moshi reports that
training Mimi with only adversarial and feature losses, dropping EnCodec's reconstruction losses,
improves subjective audio quality \citep{defossez2024moshi}. Mimi also distills
semantic information from a self-supervised speech model into its first
quantizer level while staying causal, which is the property our frame-by-frame encoding relies
on (Section~\ref{sec:overview}). We use it as a frozen, pretrained component, so no audio
tokenizer is learned as part of the adaptation.

\paragraph{Turn-taking, backchannel, and benchmarks.} Turn-taking can be modeled as the
dynamics of the two speakers' voice activity: Voice Activity Projection \citep{ekstedt2022vap}
trains a model self-supervised to predict upcoming voice activity and uses it, zero-shot, to
anticipate turn shifts and backchannels. Among the dialogue systems above, turn-taking is either
implicit, an emergent property of modeling both audio streams (dGSLM, Moshi), or explicit, as
control tokens that drive a state machine, as in \citet{wang2024fullduplex}, whose LLM decides
between responding, waiting, and interrupting. Our state machine
(Section~\ref{sec:thinkerseq}) is also explicit. The difference is granularity: its tokens share
the language model's vocabulary and output stream and are decided at every 80\,ms frame, down to
word-level pauses, under a legality mask applied before sampling. Benchmarks of these behaviors
have appeared recently. Full-Duplex-Bench \citep{lin2025fdb} scores pause handling, backchannel,
smooth turn-taking, and user interruption from the model's audio output, and its v1.5
extension \citep{lin2025fdb15} adds overlap scenarios with behavior and latency metrics.
InteractionBench \citep{averalabs2026interactionbench} drives seven systems through the same
recorded user audio and scores conversational timing, spoken-task accuracy, whisper and volume
control, and groundedness over long conversations. We evaluate on Full-Duplex-Bench v1.0 and
v1.5 and on InteractionBench.

\section{Model}
\label{sec:model}

\subsection{System overview}
\label{sec:overview}

\begin{figure}[htbp]
\centering
\includegraphics[width=\linewidth]{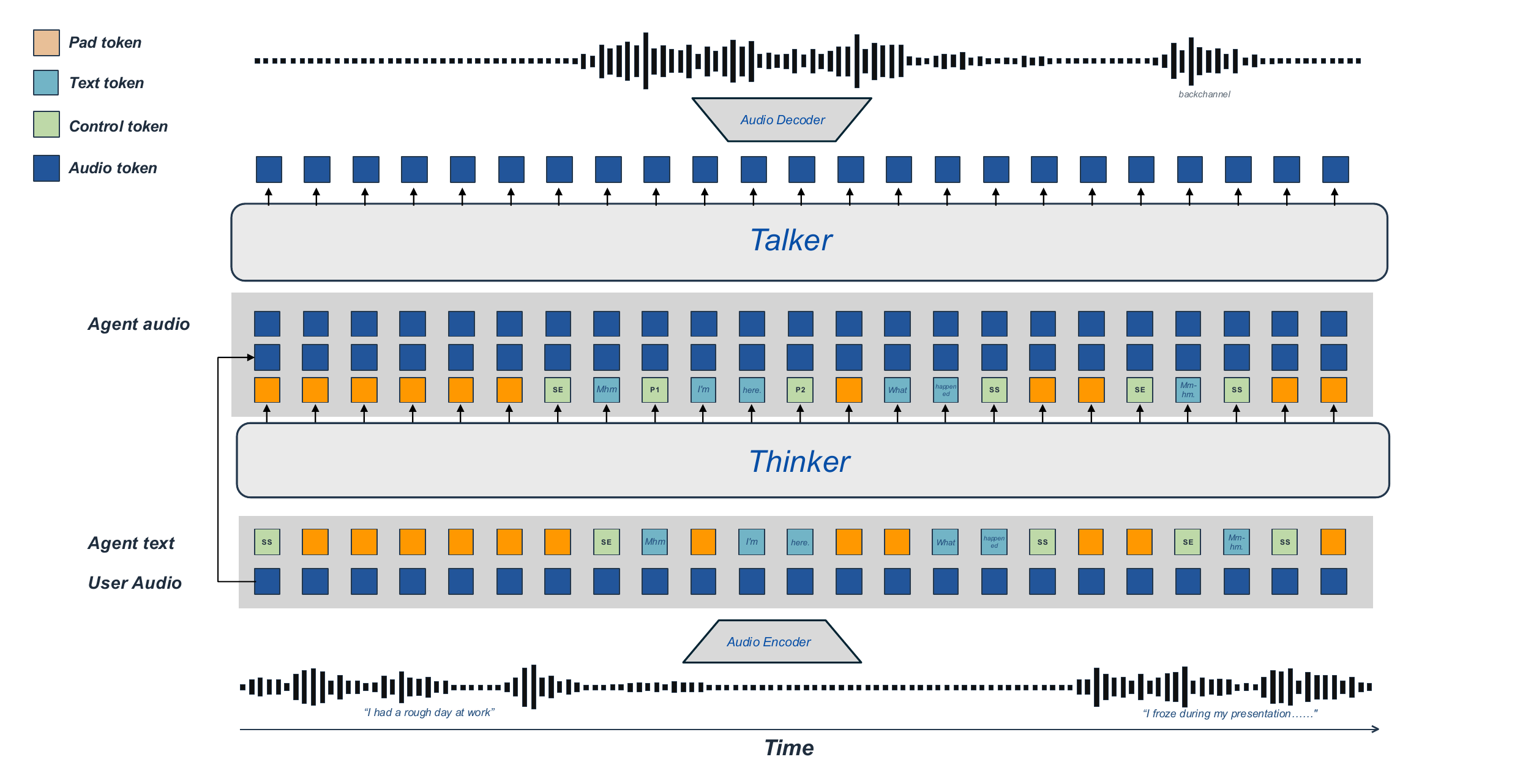}
\caption{Data flow and frame-level token structure. A frozen codec encoder turns the user's
waveform into per-frame audio tokens. The thinker reads the user-audio stream together with its
own agent-text stream (pad, text, and control tokens on one lane per frame) and extends the
text stream by one token per frame. The talker reads the agent-text stream and its own previous
audio tokens and extends the agent-audio stream by one frame; a frozen codec decoder renders it
as a waveform. The control tokens \texttt{SS} and \texttt{SE} open and close the agent's
silences, and pause tokens mark short gaps; the example includes a short backchannel (``Mhm'')
produced while the user is still speaking.}
\label{fig:dataflow}
\end{figure}

The system is a thinker--talker cascade that runs on a fixed frame clock of 12.5\,Hz (80\,ms per
frame) instead of a turn-based request--response loop (Figure~\ref{fig:dataflow}). At every
frame the thinker reads the newest slice of the user's audio and emits the next token of the
\textbf{agent-text stream}: either a word piece or a timing token that marks a pause, the start
of a silence (\texttt{SS}, the agent yields the floor), or its end (\texttt{SE}, the agent takes
the floor). The talker reads that text stream and emits the next frame of agent audio as codec
tokens, which a codec decoder turns into a waveform. Both halves advance every frame whether
or not anyone is speaking, which is what lets the agent be interrupted and backchannel.

\paragraph{Real-time budget.}
\label{sec:framebudget}
This design imposes one shared real-time requirement, which we state as a design constraint
rather than a measured result: each 80\,ms of audio, including the thinker's step and the
talker's step for that frame, must be processed within 80\,ms of wall-clock time, or frames
queue up and latency grows without bound. The budget is shared, so two halves that are each
fast enough alone are not necessarily fast enough together. Three design consequences follow.
Each audio adapter must encode only the newly arrived frame rather than re-encoding the
conversation, so that per-frame cost does not grow with turn length. Both halves must run a
forward pass during silence, since stopping the clock would reduce the system to turn-based
behavior and remove barge-in. And both must decode incrementally with a key--value cache, since
recomputing the whole prefix at every frame is quadratic in its length.

The two halves communicate only through discrete tokens. They share the codec, but each embeds
codec tokens with its own audio adapter (Section~\ref{sec:audiotok}), and the talker receives the
thinker's output as token identifiers, which it embeds with its own copy of the thinker's
token-embedding table. No continuous hidden state crosses from one to the other. As a result the
two halves share no trained parameters and are trained independently, in either order: the
talker is trained on speech paired with its word-level transcript
(Section~\ref{sec:talkertraining}), not on thinker output.

\subsection{Audio tokenization}
\label{sec:audiotok}

Neither half is trained on raw audio. The pretrained Mimi codec \citep{defossez2024moshi}
maps a waveform, in both directions, to 8 residual codebook indices per frame, $q_t = (q_t^{(0)},
\ldots, q_t^{(7)})$, coarsest first; it is kept frozen and is the only pretrained module the two
halves share at run time. Each side embeds codec tokens with its own audio adapter,
\begin{equation}
\mathcal{F}(q_t) = W\Big(\textstyle\sum_{k=0}^{7} E_k\big(q_t^{(k)}\big)\Big),
\label{eq:audiotok}
\end{equation}
where $E_k$ is the embedding table of codebook $k$ and $W$ an optional linear bridge. The
thinker's audio adapter, written $\mathcal{F}_\phi$ with parameters $\phi$, includes the bridge,
which projects the summed embedding to the thinker's hidden size; the talker's omits it, because
its summed embedding already has the talker's width. Neither audio adapter looks at other frames, so
encoding is incremental by construction. The two audio adapters have the same form but separate,
independently trained weights.

\subsection{Thinker: a finite-state model of turn-taking}
\label{sec:thinkerseq}

The thinker extends a single stream with one slot per frame, one agent-text token $y_t$ at a
time, conditioned on the user's audio on the same timeline. Its backbone is the thinker of
Qwen3-Omni-30B-A3B \citep{xu2025qwen3omni}, a mixture-of-experts LLM whose weights,
token-embedding table, and language-modeling head all stay frozen. The trainable part is small:
rank-16 low-rank adapters \citep{hu2021lora} on the attention projections (13.4M parameters),
the audio adapter $\mathcal{F}_\phi$ (75.5M), a channel-fusion layer (8.4M), and seventeen new
token embeddings for the timing tokens, each with a trainable input embedding and a trainable
output row; 97.4M parameters in all, against the backbone's 30B. The timing tokens are
\texttt{PAD} (no new token this frame), \texttt{EPAD} (the frame before a word starts), thirteen
pause tokens $\mathtt{pause}_0,\ldots,\mathtt{pause}_{12}$, \texttt{SS}, and \texttt{SE}. The
words themselves are written in the LLM's ordinary subword vocabulary.

\paragraph{Channel fusion.} Each frame's input fuses two channels, the user's audio at frame $t$
and the agent's previous token, before the backbone sees them:
\begin{equation}
x_t = W_{\text{proj}}\,\big[\,u_t \,\Vert\, e_{t-1}\,\big], \qquad
h_t = f_\theta(h_{<t}, x_t), \qquad
z_t = W_{\text{head}}\, h_t,
\label{eq:thinker-step}
\end{equation}
where $u_t = \mathcal{F}_\phi(q^{\mathrm{u}}_t)$ embeds the user's codec tokens,
$e_{t-1} = \mathrm{Emb}(y_{t-1})$ embeds the previous agent token (the frozen table for word
pieces, a trainable row for timing tokens), $f_\theta$ is one step of the backbone, and $z_t$ are
the agent-token logits. $W_{\text{proj}}$ starts at $[I\,|\,I]$, so the fused input is initially
the sum $u_t + e_{t-1}$; the \texttt{PAD} embedding starts at zero, so a frame in which the
agent is silent initially passes the user's audio through unchanged
(Section~\ref{sec:channelinit}). $W_{\text{head}}$ is the frozen language-modeling head with its
seventeen rows of the new tokens replaced by trainable ones. Because pauses and turn boundaries are
tokens, when the agent starts or stops talking is predicted by the same next-token objective as
what it says.

\begin{figure}[htbp]
\centering
\includegraphics[width=0.85\linewidth]{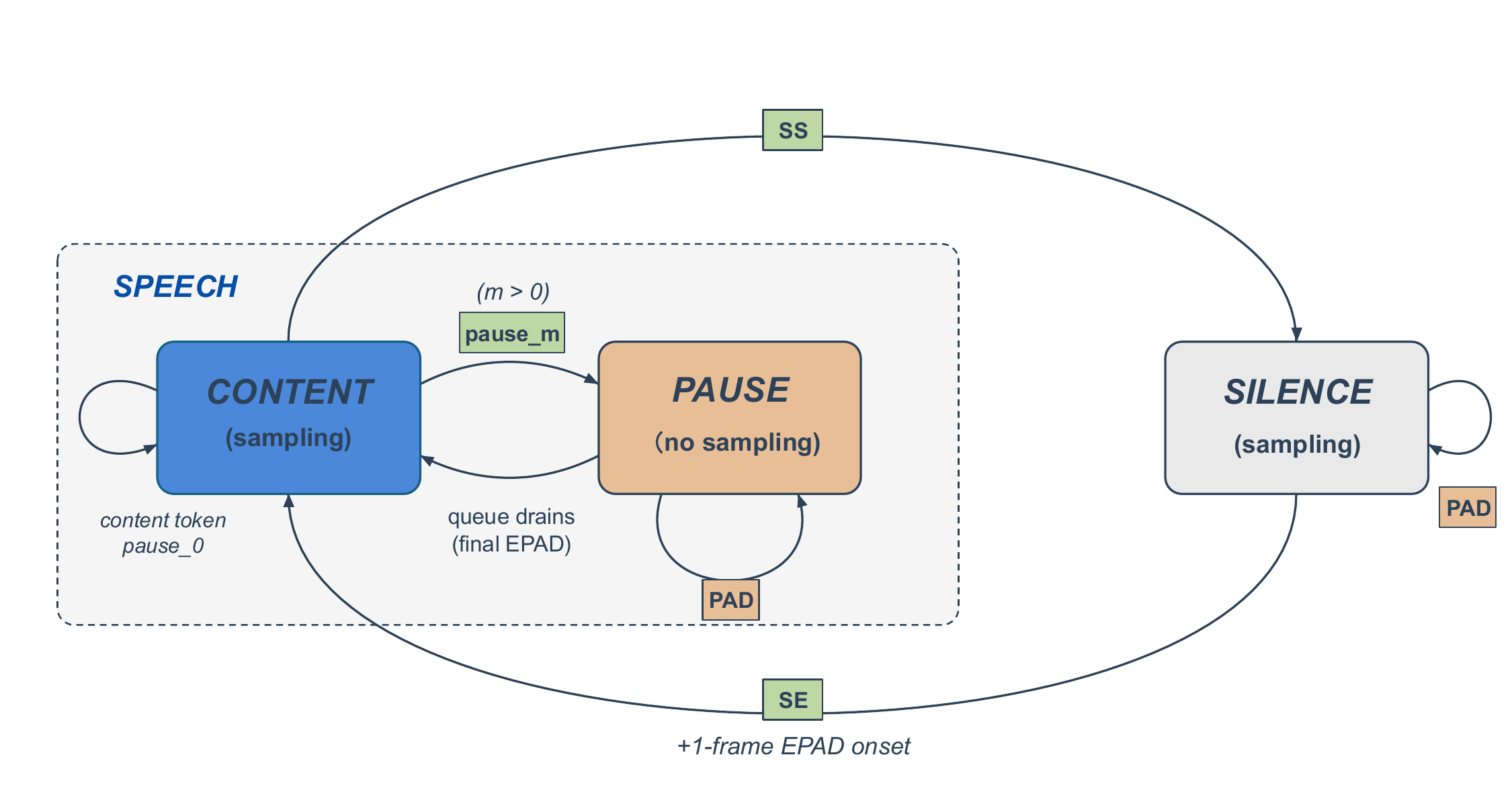}
\caption{The thinker's turn-taking state machine. In \textsc{Silence} the legal tokens are
\texttt{PAD} (stay silent) and \texttt{SE} (take the floor, entering \textsc{Content} after one
automatically fed \texttt{EPAD} frame). In \textsc{Content} they are word pieces and
$\mathtt{pause}_0$ (stay), $\mathtt{pause}_m$ with $m>0$ (enter \textsc{Pause}), and \texttt{SS}
(yield, entering \textsc{Silence}). \textsc{Pause} samples nothing: it feeds its queued
\texttt{PAD} and \texttt{EPAD} frames and returns to \textsc{Content} when the queue is empty.}
\label{fig:thinkerfsm}
\end{figure}

\paragraph{Turn-taking state machine.} Sampling at every frame is what lets the thinker decide
turn-taking word by word, and also what makes it fragile: each draw decides at once whether the
agent speaks, what it says, and whether it pauses, and a single ill-formed token can derail the
sequence. Which token may be drawn from $z_t$ is therefore decided by an explicit finite-state
machine, applied as a mask before sampling rather than as a correction afterwards. It splits the
joint decision by state: while the agent is silent, the only choice is whether to take the
floor; within its turn, the choice is the next word, a pause, or yielding; during a pause, there
is nothing to decide. Let $s_t \in \{\textsc{Content}, \textsc{Pause}, \textsc{Silence}\}$ be the state at
frame $t$. \textsc{Content} and \textsc{Pause} together form the agent's turn (the dashed
\textsc{Speech} region in Figure~\ref{fig:thinkerfsm}); \textsc{Silence} means the agent has
yielded. Sampling and transition are
\begin{equation}
p_\theta(y_t \mid h_t, s_t) = \mathrm{softmax}\big(z_t + \mathrm{mask}(s_t)\big), \qquad
\mathrm{mask}(s_t)_v = \begin{cases} 0, & v\in\mathrm{legal}(s_t), \\ -\infty, & \text{otherwise,} \end{cases}
\label{eq:thinker-mask}
\end{equation}
\begin{equation}
s_{t+1} = \delta(s_t, y_t, Q_t), \qquad
\delta = \begin{cases}
\textsc{Silence}, & s=\textsc{Content},\ y=\mathtt{SS}, \\
\textsc{Pause}, & s=\textsc{Content},\ y=\mathtt{pause}_{m},\ m>0, \\
\textsc{Content}, & s=\textsc{Content},\ y \text{ a word piece or } \mathtt{pause}_0, \\
\textsc{Silence}, & s=\textsc{Silence},\ y=\mathtt{PAD}, \\
\textsc{Content}, & s=\textsc{Silence},\ y=\mathtt{SE}, \\
\textsc{Pause}, & s=\textsc{Pause},\ |Q_t|>1, \\
\textsc{Content}, & s=\textsc{Pause},\ |Q_t|=1,
\end{cases}
\label{eq:thinker-transition}
\end{equation}
with $\mathrm{legal}(\textsc{Content}) = V \setminus \{\mathtt{PAD}, \mathtt{EPAD}, \mathtt{SE}\}$
and $\mathrm{legal}(\textsc{Silence}) = \{\mathtt{PAD}, \mathtt{SE}\}$. \textsc{Pause} is not a
sampling state. Emitting $\mathtt{pause}_m$ with $m>0$ commits $m{-}1$ \texttt{PAD} frames
followed by one \texttt{EPAD} to a first-in, first-out queue $Q_t$; while the queue is non-empty,
$y_t$ is popped from it instead of being sampled, and the frame that pops its last element
returns control to \textsc{Content}. Emitting \texttt{SE} likewise queues one \texttt{EPAD}, so
the frame before the agent's first word is fed automatically. A separate grammar checker
enforces the same rules on every generated sequence before it is scored.

The state machine also fixes how a conversation begins. In one mode the thinker emits
\texttt{SS} at frame 0 without sampling, so the agent opens by yielding and listening; in the
other, used for evaluations in which the user is known to speak first, it starts directly in
\textsc{Silence}. Either way the agent's first decision is whether to take the floor.

\subsection{Talker: streaming speech synthesis from the agent-text stream}
\label{sec:talker}

The talker also extends a frame-aligned stream, but its output is codec tokens rather than text.
Its architecture is Moshi's \citep{defossez2024moshi} at reduced scale: a temporal transformer
that advances once per frame and a depth transformer that resolves, within the frame, one
codebook after another. Unlike the thinker, the talker is fine-tuned in full rather than through
adapters, starting from a pretrained speech-text checkpoint (Section~\ref{sec:talkertraining}).

\paragraph{Streams and delays.} The talker models three streams on the frame grid: the agent
text $y$, which is given, never predicted, and carries no loss; the agent's codec tokens
$c^{(0)},\ldots,c^{(7)}$, which it predicts; and the user's codec tokens $q^{\mathrm{u}}$. Two
offsets align them. The text stream leads the audio by three frames (240\,ms), so that the
talker sees a word slightly before it must voice it; we write $\tilde y_t = y_{t+3}$. And,
following the acoustic delay pattern introduced for parallel prediction of residual codebooks
by \citet{copet2023musicgen} and adopted by Moshi, the seven residual codebooks lag the coarse
one by a frame: $\tilde c^{(0)}_t = c^{(0)}_t$ and $\tilde c^{(k)}_t = c^{(k)}_{t-1}$ for
$k \geq 1$, and the user's codebooks are laid out the same way. Each sequence is preceded
by a short voice prompt, a reference recording of the target voice with its transcript on the
text stream, which sets the voice the talker speaks in (Section~\ref{sec:talkertraining}).

\paragraph{Channel fusion.} Following Moshi's input design, the temporal transformer's input at
step $t$ is the sum of the embeddings of every stream at step $t-1$:
\begin{equation}
v_t = P\,\mathrm{Emb}'(\tilde y_{t-1}) + \sum_{k=0}^{7} A_k\big(\tilde c^{(k)}_{t-1}\big) +
\sum_{k=0}^{7} U_k\big(\tilde q^{\mathrm{u},(k)}_{t-1}\big), \qquad
\bar h_t = f_\psi(\bar h_{<t}, v_t),
\label{eq:talker-fuse}
\end{equation}
where $\mathrm{Emb}'$ is the talker's copy of the thinker's token-embedding table, $P$ a trainable
projection to the talker's width, $A_k$ and $U_k$ the per-codebook embedding tables of the agent
and user streams (the talker's audio adapter, Equation~\ref{eq:audiotok} without a bridge), and
$f_\psi$ one step of the temporal transformer. Summation is also where the thinker's channel fusion
starts (Section~\ref{sec:channelinit}). In the talker evaluated in this paper the user-stream
embeddings $U_k$ are held at zero, so the talker is a streaming synthesizer conditioned on the
thinker's text; conditioning the voice on the user's audio is supported by the design but not
exercised (Section~\ref{sec:limitations}).

\paragraph{Hierarchical codebook prediction.} Following the RQ-Transformer decomposition
\citep{lee2022rqtransformer} as adopted by Moshi, the depth transformer $D_\psi$ predicts the
step's eight codebooks in sequence rather than independently, each conditioned on the temporal
state and on the codebooks already chosen:
\begin{equation}
\begin{gathered}
\tilde c^{(k)}_t \sim \mathrm{softmax}\Big(L_k\, D_\psi\big(d_0, \ldots, d_k\big)_k\Big),\\
d_0 = G_0\,\bar h_t + \mathrm{Emb}''(\tilde y_t), \qquad
d_k = G_k\,\bar h_t + A'_{k}\big(\tilde c^{(k-1)}_t\big)\ \ (k \geq 1),
\end{gathered}
\label{eq:talker-mtp}
\end{equation}
for $k = 0, \ldots, 7$, where $G_k$ are per-codebook projections of the temporal state, $L_k$
per-codebook output layers, $\mathrm{Emb}''$ the depth transformer's own text embedding (also a
copy of the thinker's table with a projection), and $A'_k$ its codebook embeddings. Because of
the delay, one step emits the coarse codebook of frame $t$ together with the residual codebooks
of frame $t-1$, which are therefore chosen with one frame of look-ahead from the coarse
codebook that follows them; a frame's eight codes are complete one step after its coarse code.
The silence markers reach the talker through its text stream: an \texttt{SS} or \texttt{SE}
token in $\tilde y$ tells it when the agent's silence begins or ends.

\paragraph{Inference.} The talker runs Equations~\ref{eq:talker-fuse} and~\ref{eq:talker-mtp} once
per frame, with the thinker's tokens in place of the transcript. The coarse codebook is sampled
at temperature 0.8; lower temperatures were observed to collapse into silence on a fraction of
conversations (Appendix~\ref{app:impl}). Completed frames are fed back as agent-audio input. The
serving path decodes them incrementally with Mimi's streaming decoder, a few frames at a time;
the samples used for evaluation are decoded offline from the complete code sequence.

\section{Training}
\label{sec:training}

Neither half of the cascade starts from a full-duplex model. The thinker starts from a
general-purpose multimodal LLM and the talker from a pretrained speech-text model, and neither
has any notion of a frame-aligned, two-channel dialogue. This section describes how each is moved
onto that format. Section~\ref{sec:channelinit} gives the initialization rule that lets new input
paths join a frozen LLM without disturbing it, Section~\ref{sec:curriculum} gives the thinker's
four-stage curriculum, and Section~\ref{sec:talkertraining} gives the talker's training, which
is a single supervised stage. Table~\ref{tab:curriculum} summarizes what each training run
trains, on how much data, and at what cost.

\subsection{Channel initialization}
\label{sec:channelinit}

Every new input path joins the pretrained model by addition, and is initialized so that, on the
first step of training, the model receives an input it already knows how to read. In the
thinker, the channel-fusion layer of Equation~\ref{eq:thinker-step} starts at $W_{\text{proj}} = [I\,|\,I]$,
so the fused input is $u_t + e_{t-1}$: the LLM's own token embedding plus an audio embedding that
Stage~1 trained to be readable by the frozen LLM. Because the \texttt{PAD} embedding also starts
at zero, a frame in which the agent is silent reaches the backbone as the user's audio alone. In
the talker the streams are summed by construction (Equation~\ref{eq:talker-fuse}). This choice
rules out two alternatives. A randomly initialized channel-fusion layer would perturb the pretrained
model's input at the start of training, and with it the knowledge the conversion is meant to
keep. A block initialized at zero starts with its channel switched off, and nothing forces
training to switch it back on: in one earlier run of the thinker, a channel-fusion layer initialized as $[0\,|\,I]$
ended training with a user-audio block of weight norm about 1.2, against about 64 with $[I\,|\,I]$,
and could not recall facts that had appeared only in the user's speech.

\subsection{Training curriculum}
\label{sec:curriculum}

The thinker's backbone, token embeddings, and language-modeling head stay frozen throughout.
The curriculum has four stages; each trains a small set of added parameters
(Section~\ref{sec:thinkerseq}), closes one gap left by the stage before it, and is initialized
from its predecessor. Stages~1, 3, and~4 each consist of two training runs, described below. Figure~\ref{fig:overview} shows, for each stage, what is trained, what the
model reads and emits, and which model anchors the KL term; Table~\ref{tab:curriculum} gives
the data and compute.

\paragraph{Stage 1: reading codec audio.} Only the audio adapter $\mathcal{F}_\phi$ of
Equation~\ref{eq:audiotok} is trained; the LLM's parameters $\theta$ do not change. The audio adapter
is first trained as an ASR read-out: prompted to transcribe, the frozen LLM must reproduce the
transcript $\ell$ of the speech whose codec tokens $q$ it receives through the audio adapter,
\begin{equation}
\phi_0 = \arg\min_\phi\; \mathbb{E}_{(q,\ell)}\Big[-\log p_\theta\big(\ell \mid
\pi_{\text{pre}},\, \mathcal{F}_\phi(q),\, \pi_{\text{post}}\big)\Big],
\label{eq:stage1}
\end{equation}
where $\pi_{\text{pre}}$ and $\pi_{\text{post}}$ are fixed text prompts. The audio adapter is then
distilled against the same frozen LLM listening through its own native audio encoder. For a
spoken question with waveform $\omega$ and codec tokens $q$, the teacher is the LLM given $\omega$
through that encoder, $\mathrm{Enc}(\omega)$, and the student is the same LLM given
$\mathcal{F}_\phi(q)$, both under the same prompt; the two are matched on the positions
$\mathcal{A}$ of the teacher's own answer $r$,
\begin{equation}
\phi^\star = \arg\min_\phi\; \mathbb{E}_{(\omega,q)} \sum_{j\in\mathcal{A}} \tau^2\,
\mathrm{KL}\Big(p^{\tau}_\theta\big(\cdot \mid \mathrm{Enc}(\omega), r_{<j}\big) \,\Big\Vert\,
p^{\tau}_\theta\big(\cdot \mid \mathcal{F}_\phi(q), r_{<j}\big)\Big),
\label{eq:stage1-distill}
\end{equation}
where $p^{\tau}$ is the output distribution at temperature $\tau = 1.5$, and $\phi$ starts from
$\phi_0$. The read-out step trains on 17{,}210 windows of 30 to 90 seconds of user speech cut
from 8{,}000 dialogues (about 270 hours), and the distillation step on 3{,}911 spoken questions
(6.3 hours). The LLM already understands speech through its own encoder; this stage only
teaches it to read the codec tokens the rest of the system uses, which is why the LLM itself
does not change.

\paragraph{Stage 2: pause timing.} With the audio adapter frozen, the model learns the vocabulary for
timing: \texttt{PAD}, \texttt{EPAD}, and thirteen pause-duration tokens occupy fifteen reserved
rows, trained together with rank-16 low-rank adapters on 760 single-turn spoken questions whose
answers carry word-level pauses, with an ordinary cross-entropy loss. This stage teaches where
the agent's own speech has gaps before the next stage teaches whose turn it is.

From Stage~3 onward every stage uses the same form of objective. With $\kappa(y_t)$ the class of
target token $y_t$ (word piece, pause, \texttt{PAD}, \texttt{EPAD}, \texttt{SS}, \texttt{SE}),
$\mathcal{C}$ the content positions, and $p_{\text{ref}}$ a frozen reference model,
\begin{equation}
\mathcal{L}(\theta') = \sum_t \lambda_{\kappa(y_t)}\big[-\log p_{\theta'}(y_t \mid
\cdot)\big] \;+\; \beta \sum_{t\in\mathcal{C}}
\mathrm{KL}\big(p_{\text{ref}}(\cdot\mid\cdot) \,\Vert\, p_{\theta'}(\cdot\mid\cdot)\big).
\label{eq:stage-kl}
\end{equation}
The class weights $\lambda$ down-weight \texttt{PAD} (0.3), which fills most agent frames, and
weight the turn events \texttt{SS} and \texttt{SE} at 1.5; word pieces are weighted 1.5 in
the first run of Stage~3 and 1.0 afterwards. The timing tokens are learned by the cross-entropy term. The KL
term applies only at word positions, and holds what the model says close to a reference that
already says it well. In every stage the reference is the model the stage was initialized from.

\paragraph{Stage 3: from one channel to two.} This stage moves the model from a single-channel
language model onto the dual-channel format of Section~\ref{sec:thinkerseq}, adding the turn
events \texttt{SS} and \texttt{SE} (seventeen new token embeddings in total), in two runs. The first
is self-distillation on 760 examples: the reference $p_{\text{ref}}$ is the
frozen Stage-2 model reading the same example in its native single-stream format, and the
student reads it in the new dual-channel format, with $\beta = 0.8$ and the channel-fusion layer held at
$[I\,|\,I]$. The student's word choices are thus anchored to its own behavior under the old
format while it learns where the new timing tokens go. The user channel outside the question is
a constant digital-silence template. Stage~3 then ends with a multi-turn retrain under natural
room tone: 12{,}146 dialogues (406 hours) whose replies were generated offline by the
frozen LLM itself, sampled several times per question, in which the user channel carries a
continuous room-tone floor at a random level between $-75$ and $-45$\,dBFS instead of digital
silence. The channel-fusion layer and the audio adapter are trained from this run on, for one epoch with
$\beta = 0.4$. At this point the model takes turns in multi-turn dialogue, but its training data
still contains no overlap between the two speakers.

\paragraph{Stage 4: overlap, interruption, and backchannel.} The last stage teaches behavior that
is not turn-based at all, in two runs that share the objective of Equation~\ref{eq:stage-kl}
with $\beta = 0.4$ and train the channel-fusion layer and the audio adapter. The first run teaches the
model to yield when interrupted and to understand speech that overlaps its own. Its 36{,}621
dialogues (1{,}058 hours) are the Stage-3 dialogues with interruptions injected: with
probability 0.35 per agent answer, the user's next question is moved earlier so that it begins
after 35 to 70\% of the answer and overlaps it for 0.8 to 1.3 seconds, after which the answer is
cut and the agent must yield; 83\% of the dialogues contain at least one such interruption. The
user channel carries a room-tone floor between $-78$ and $-63$\,dBFS. Training runs for 3{,}000
steps, just under one epoch. The second run adds backchannel. Its 42{,}229 sequences are the
barge-in dialogues (86.7\%) and 701 dialogues in which the agent backchannels while the user
speaks, each repeated eight times (13.3\%), under a single persona prompt, for 3{,}000 steps
(0.85 epochs). This run completes the curriculum.

\begin{table}[t]
\centering
\caption{The thinker's conversion curriculum. Parameters not listed stay frozen, including the
entire LLM, its token embeddings, and its language-modeling head. ``Embeddings'' are the new token
embeddings for the timing tokens. ``Reference'' is the model that anchors the KL term of
Equation~\ref{eq:stage-kl}. Hours are the duration of the user-channel timeline of the training
data; GPU-hours are measured on one NVIDIA B200 and exclude data generation and evaluation.
One row per training run, grouped by stage.}
\label{tab:curriculum}
\resizebox{\linewidth}{!}{
\begin{tabular}{@{}llrrrllr@{}}
\toprule
Stage, run & Gap closed & Examples & Hours & Steps & Trained & Reference ($\beta$) & GPU-h \\
\midrule
1, read-out & read codec audio & 17{,}210 & $\approx$270 & 2{,}000 & audio adapter & -- & 2.6 \\
1, distillation & & 3{,}911 & 6.3 & 733 & audio adapter & native-audio LLM & 0.5 \\
2 & pause timing & 760 & 1.0 & 285 & adapters, 15 embeddings & -- & 1.2 \\
3, self-distillation & dual-channel format & 760 & 2.1 & 285 & adapters, 17 embeddings & Stage 2 (0.8) & 0.9 \\
3, multi-turn retrain & room tone, multi-turn & 12{,}146 & 406 & 1{,}012 & + channel fusion, audio adapter & previous run (0.4) & 3.2 \\
4, barge-in & interruption & 36{,}621 & 1{,}058 & 3{,}000 & + channel fusion, audio adapter & Stage 3 (0.4) & 9.1 \\
4, backchannel & backchannel & 42{,}229 & 1{,}150 & 3{,}000 & + channel fusion, audio adapter & previous run (0.4) & 13.0 \\
\midrule
\multicolumn{7}{@{}l}{Total} & 30.5 \\
\bottomrule
\end{tabular}}
\end{table}

\subsection{Talker training}
\label{sec:talkertraining}

The talker is trained in a single supervised stage. It is initialized from an earlier talker
checkpoint whose lineage begins from PersonaPlex \citep{roy2026personaplex}, a full-duplex model
fine-tuned from Moshi. Each training example is a segment of synthesized dialogue speech together
with its word-level transcript, whose alignment is computed upstream of this system and stored
with the audio. The transcript is written on the agent-text stream in exactly the format the
thinker emits (silence-start and silence-end tokens, and pauses rendered as \texttt{PAD}), so the
talker is trained on the input it receives at run time, and every training sequence passes the
grammar checker of Section~\ref{sec:thinkerseq}. The text leads the audio by three frames, and the
speech is encoded as the agent-audio stream. Because the transcript stands in for the thinker's
output, no thinker is needed to produce training data.

The objective is the mean cross-entropy of the eight agent codebooks under the factorization of
Equation~\ref{eq:talker-mtp}, with teacher forcing on the coarser codebooks:
\begin{equation}
\mathcal{L}_{\text{talker}}(\psi) = \frac{1}{8\,|\mathcal{M}|} \sum_{t \in \mathcal{M}}
\sum_{k=0}^{7} -\log p_\psi\big(\tilde c^{(k)}_t \,\big|\, \bar h_t,\, \tilde c^{(0)}_t, \ldots,
\tilde c^{(k-1)}_t\big),
\label{eq:talker-loss}
\end{equation}
where $\bar h_t$ depends on all three streams up to step $t-1$ through Equation~\ref{eq:talker-fuse},
and $\mathcal{M}$ is the set of scored frames: the dialogue frames, excluding the voice-prompt
prefix and its spacer, which are conditioning only, and excluding any voiced but wordless
stretch before the first transcribed word, so that the talker is not trained to vocalize without
text. The text stream and the user stream carry no loss, and all codebooks and frames are
weighted equally.

Each sequence begins with a voice prompt: 60 frames (4.8\,s) of a reference recording of the
target voice, with that recording's frame-aligned transcript on the text stream, followed by
12 silent frames (0.96\,s) and then the dialogue. PersonaPlex \citep{roy2026personaplex} also
conditions its voice on a short reference recording. At inference the same kind of prompt sets
the agent's voice; by default it is a single voice from a commercial text-to-speech service.

The training set has 36{,}996 examples, kept from 60{,}926 by four filters that remove
(i)~examples whose audio length does not match their transcript grid, a known generation fault in
part of the synthetic data; (ii)~examples whose agent audio has a speech-recognition word error
rate above 0.15 against its own transcript, close to the 95th percentile measured on clean
text-to-speech audio; (iii)~examples with audio energy before the first word; and
(iv)~examples whose voice prompt is mostly silent, carries fewer than three words, or is too
quiet. Sequences are truncated at 96\,s. Training runs for four epochs (35{,}504 steps at
batch size 4) at a learning rate of $10^{-5}$ with cosine decay and 5\% warmup, and takes 2.65
hours on one NVIDIA B200. Checkpoints were compared by the speech-recognition word error rate of
their output on held-out samples (Section~\ref{sec:talkerwer}); we use the final checkpoint of
the four-epoch run, since four further epochs did not change the held-out error rate (paired
difference $-0.005$, 95\% confidence interval $[-0.015, 0.004]$, over 150 samples and three
seeds).

\section{Experimental Setup}
\label{sec:setup}

\subsection{Benchmarks}
\label{sec:benchmarks}

\paragraph{Full-Duplex-Bench v1.0} \citep{lin2025fdb} measures four turn-taking behaviors from
recorded or synthesized user audio: pause handling (the model should not take the turn when the
user merely pauses), backchannel, smooth turn-taking (the model should take the turn when the
user finishes), and user interruption. The benchmark reports the turn-over rate (TOR), the
fraction of samples in which the model takes the turn; lower is better for pause handling and
backchannel, higher for smooth turn-taking and user interruption. For backchannel it also reports
frequency, in backchannels per second of audio, and the Jensen--Shannon divergence between the
timing of the model's backchannels over each clip and a human reference distribution. Latencies
are in seconds. After a user interruption, an LLM judge rates the relevance of the model's response
from 0 to 5. Each run scores 353 pause-handling samples (216 recorded, 137 synthetic), 55
backchannel samples, 119 turn-taking samples, and 200 interruption samples.

\paragraph{Full-Duplex-Bench v1.5} \citep{lin2025fdb15} adds four overlap scenarios: the user
interrupts, the user backchannels, a third party talks to the user, and background speech plays
while the model is speaking. For each, the model's speech after the overlap is classified as a
response to the overlap, a resumption of what it was saying, uncertain, or unknown, and the
benchmark reports the stop latency (how quickly the model stops after the overlap begins) and the
response latency (how quickly it starts a new response), in seconds.

\paragraph{InteractionBench} \citep{averalabs2026interactionbench} drives every system through
the same recorded user audio and scores four dimensions in 221 items over thirteen scenarios:
conversational timing (backchannel, pause, turn-taking, user backchannel, and interruption
scenarios with fixed pass criteria), spoken-task accuracy (50 items in which the expected
floor-control action and answer are fixed in advance), whisper and volume control (20 items),
and groundedness (52 probes asked inside ten conversations of more than twenty turns, plus a
coherence score from 0 to 100). It publishes results for seven systems.

\subsection{Baselines}
\label{sec:baselines}

Every comparison table lists the systems in two blocks: open-source models that were
self-hosted, and closed-source systems reached through an API. The comparisons in the text and
the bold marks in the tables are restricted to the open-source block; the closed-source rows are
reproduced for reference only. On Full-Duplex-Bench the open-source systems are PersonaPlex,
dGSLM, Moshi, and Freeze-Omni, and the closed-source systems are Gemini Live and GPT-Realtime
(GPT-4o Realtime on v1.5); their numbers are taken from the benchmarks' published results. On InteractionBench the open-source systems are Moshi, PersonaPlex, and
Freeze-Omni, and the closed-source systems are GPT-Realtime and Gemini Live. All comparisons with
other systems use these public benchmarks; the noise-floor analysis in Section~\ref{sec:analysis}
uses a small internal question-answering probe.

\subsection{Protocol}
\label{sec:protocol}

The full system runs end to end: the thinker produces the agent-text stream from the user audio,
the talker renders it as speech, and each benchmark's own transcription and evaluation scripts
score the result without modification. The output audio has the same length as the input, as
the benchmarks expect. No external voice-activity detector is used, so the model decides every
turn itself. The thinker runs with the persona system prompt it was trained with, and every row
is voiced by the same talker (Section~\ref{sec:talkertraining}) sampling at temperature 0.8. We
report means and sample standard deviations over three decoding seeds on Full-Duplex-Bench; the
InteractionBench row is a single decoding seed (Section~\ref{sec:ibresults}).

\section{Results}
\label{sec:results}

\subsection{Full-Duplex-Bench v1.0}
\label{sec:fdbresults}

Table~\ref{tab:fdb} compares our system with the published results of the open-source systems;
the closed-source rows are reproduced for reference and are not part of the comparison. The
clearest difference is in pause handling: our system is the least likely of the open-source
systems to take the floor when the user merely pauses, on both subsets and by a wide margin
(turn-over rate 0.170 on the synthetic subset against 0.584 for the next-best system). It also
has the lowest turn-over rate on the backchannel task and the highest backchannel frequency, and
it takes the turn in nearly every smooth turn-taking case. Its weaknesses are responsiveness and
content: PersonaPlex and Moshi respond faster, both when the user finishes a turn and after an
interruption; the timing of PersonaPlex's backchannels is closer to the human distribution; and
after an interruption our responses are rated less relevant than those of PersonaPlex and
Freeze-Omni.

\begin{table}[t]
\centering
\caption{Full-Duplex-Bench v1.0. Baseline results are from the benchmark's public results
table \citep{lin2025fdb}. Arrows give the better direction. We
compare with the open-source systems: bold marks the best value in that block, and the
closed-source, API-based rows are for reference only. Relevance is rated 0--5 by the benchmark's
LLM judge (GPT-4-turbo in its released evaluation script), run unmodified, as for the other rows. Latencies are in seconds. Our row is the mean
over three seeds; standard deviations are in Appendix~\ref{app:tables}.}
\label{tab:fdb}
\resizebox{\linewidth}{!}{
\begin{tabular}{@{}lcccccccccc@{}}
\toprule
& \multicolumn{2}{c}{Pause handling} & \multicolumn{3}{c}{Backchannel} &
\multicolumn{2}{c}{Smooth turn-taking} & \multicolumn{3}{c}{User interruption} \\
\cmidrule(lr){2-3}\cmidrule(lr){4-6}\cmidrule(lr){7-8}\cmidrule(lr){9-11}
Model & Synth.\ TOR$\downarrow$ & Candor TOR$\downarrow$ & TOR$\downarrow$ & Freq$\uparrow$ &
JSD$\downarrow$ & TOR$\uparrow$ & Lat.$\downarrow$ & TOR$\uparrow$ & Relev.$\uparrow$ &
Lat.$\downarrow$ \\
\midrule
\multicolumn{11}{@{}l}{\textit{Open-source (self-hosted)}} \\
PersonaPlex & 0.584 & 0.662 & 0.327 & 0.025 & \textbf{0.649} & 0.992 & \textbf{0.070} & \textbf{1.000} & \textbf{4.210} & 0.400 \\
dGSLM & 0.934 & 0.935 & 0.691 & 0.015 & 0.934 & 0.975 & 0.352 & 0.917 & 0.201 & 2.531 \\
Moshi & 0.985 & 0.980 & 1.000 & 0.001 & 0.957 & 0.941 & 0.265 & \textbf{1.000} & 0.765 & \textbf{0.257} \\
Freeze-Omni & 0.642 & 0.481 & 0.636 & 0.001 & 0.997 & 0.336 & 0.953 & 0.867 & 3.615 & 1.409 \\
Ours & \textbf{0.170} & \textbf{0.267} & \textbf{0.145} & \textbf{0.073} & 0.753 & \textbf{0.997} & 0.44 & 0.998 & 2.681 & 0.598 \\
\midrule
\multicolumn{11}{@{}l}{\textit{Closed-source / API-based (reference)}} \\
Gemini Live 2.0 & 0.255 & 0.310 & 0.091 & 0.012 & 0.896 & 0.655 & 1.301 & 0.891 & 3.376 & 1.183 \\
GPT-Realtime & 0.010 & 0.120 & 0.000 & 0.007 & 0.980 & 1.000 & 1.470 & 0.970 & 3.850 & 1.500 \\
\bottomrule
\end{tabular}}
\end{table}

\subsection{Full-Duplex-Bench v1.5}
\label{sec:fdb15results}

Table~\ref{tab:fdb15} reports behavior after each kind of overlap, again compared with the
open-source systems. Our system reacts fastest: after a user interruption it stops talking and
starts its response sooner than Freeze-Omni and Moshi, and its response latency is the lowest in
every scenario. It also responds least often to speech that is not addressed to it, after a
side conversation or background speech, and, tied with Moshi, after a user backchannel. Two
weaknesses remain. After an interruption it responds less often than Freeze-Omni and Moshi, and
after a user backchannel it resumes its earlier utterance far less often than Freeze-Omni (0.24
against 0.80); Freeze-Omni also keeps talking longer through a side conversation or background
speech.

\begin{table}[t]
\centering
\caption{Full-Duplex-Bench v1.5. For each overlap scenario, the share of the model's
post-overlap speech classified as a response to the overlap (Resp.) or a resumption of its
earlier utterance (Res.), and the stop and response latencies in seconds after the overlap
begins. Baseline results are from \citet[Tables 2 and 3]{lin2025fdb15}; the uncertain and
unknown behavior classes are omitted. Which behavior is desirable depends on the
scenario: a model should respond to the user's interruption but keep going after a backchannel,
a side conversation, or background speech, so the arrows mark the desirable direction in each
scenario (stop late, if at all, in the keep-going scenarios; respond quickly whenever a response
is given). We compare with the open-source systems: bold marks the best value in that block,
and the closed-source, API-based rows are for reference only. Our row is the mean over three
seeds; standard deviations are in Appendix~\ref{app:tables}.}
\label{tab:fdb15}
\resizebox{\linewidth}{!}{
\begin{tabular}{@{}lcccccccccccc@{}}
\toprule
& \multicolumn{3}{c}{User interruption} & \multicolumn{3}{c}{User backchannel} &
\multicolumn{3}{c}{Side conversation} & \multicolumn{3}{c}{Background speech} \\
\cmidrule(lr){2-4}\cmidrule(lr){5-7}\cmidrule(lr){8-10}\cmidrule(lr){11-13}
Model & Resp.$\uparrow$ & Res.$\downarrow$ & Stop$\downarrow$ / Resp.$\downarrow$ lat. & Resp.$\downarrow$ & Res.$\uparrow$ & Stop$\uparrow$ / Resp.$\downarrow$ lat. & Resp.$\downarrow$ & Res.$\uparrow$ & Stop$\uparrow$ / Resp.$\downarrow$ lat. & Resp.$\downarrow$ & Res.$\uparrow$ & Stop$\uparrow$ / Resp.$\downarrow$ lat. \\
\midrule
\multicolumn{13}{@{}l}{\textit{Open-source (self-hosted)}} \\
Freeze-Omni & \textbf{0.72} & 0.12 & 1.42 / 1.35 & 0.07 & \textbf{0.80} & 0.66 / 2.16 & 0.58 & \textbf{0.25} & \textbf{1.39} / 1.32 & 0.63 & \textbf{0.25} & \textbf{0.98} / 1.60 \\
Moshi & 0.50 & 0.26 & 1.16 / 1.47 & \textbf{0.02} & 0.06 & 0.42 / 3.00 & 0.20 & 0.19 & 0.87 / 2.38 & 0.21 & 0.07 & 0.54 / 1.62 \\
Ours & 0.42 & \textbf{0.10} & \textbf{0.75} / \textbf{0.64} & \textbf{0.02} & 0.24 & \textbf{0.68} / \textbf{1.33} & \textbf{0.08} & \textbf{0.25} & 0.84 / \textbf{0.62} & \textbf{0.06} & 0.23 & 0.83 / \textbf{0.64} \\
\midrule
\multicolumn{13}{@{}l}{\textit{Closed-source / API-based (reference)}} \\
Gemini Live & 0.33 & 0.55 & 2.20 / 2.62 & 0.01 & 0.93 & 0.66 / 2.45 & 0.00 & 0.99 & 1.69 / 1.78 & 0.70 & 0.30 & 0.95 / 2.38 \\
GPT-4o Realtime & 0.78 & 0.10 & 0.23 / 1.50 & 0.03 & 0.70 & 0.21 / 1.32 & 0.91 & 0.02 & 0.18 / 1.16 & 0.93 & 0.04 & 0.18 / 1.26 \\
\bottomrule
\end{tabular}}
\end{table}

\subsection{InteractionBench}
\label{sec:ibresults}

Table~\ref{tab:ib} reports the four dimensions, compared with the three open-source systems
published with the benchmark. Among them, ours is the strongest on groundedness, coherence, and
spoken tasks, with groundedness well above the next-best system (53\% against 35\% for
Freeze-Omni), and second on conversational timing, behind PersonaPlex. It is the weakest on
whisper and volume, which our talker does not control (Section~\ref{sec:limitations}). The
closed-source systems remain ahead on every dimension. Table~\ref{tab:ib-scenarios} breaks our
result down by scenario.

\begin{table}[t]
\centering
\caption{InteractionBench. Percentages are over the items in each dimension (counts in
parentheses); coherence is the mean judge score over the ten groundedness conversations.
Baseline results are from \citet[Table 4]{averalabs2026interactionbench}. We compare with the open-source systems: bold marks the best value in
that block, and the closed-source, API-based rows are for reference only. Our row is a single
decoding seed; its groundedness rate is over the 51 valid probes (one probe was excluded as
invalid). Our row is graded on the thinker's text stream, with each word timed at the frame in
which the thinker emits it; the grader components that read audio use the talker's output, cut
at those times.}
\label{tab:ib}
\resizebox{\linewidth}{!}{
\begin{tabular}{@{}lccccc@{}}
\toprule
System & Spoken task (\%)$\uparrow$ & Timing (\%)$\uparrow$ & Whisper \& volume (\%)$\uparrow$ & Grounded (\%)$\uparrow$ & Coherence (0--100)$\uparrow$ \\
\midrule
\multicolumn{6}{@{}l}{\textit{Open-source (self-hosted)}} \\
Moshi & 6 (3/50) & 41 (54/132) & 15 (3/20) & 14 (7/50) & 16 \\
PersonaPlex & 18 (9/50) & \textbf{68} (90/132) & \textbf{35} (7/20) & 30 (15/50) & 28 \\
Freeze-Omni & 8 (4/50) & 57 (78/138) & 15 (3/20) & 35 (18/51) & 39 \\
Ours & \textbf{26} (13/50) & 63 (88/140) & 5 (1/20) & \textbf{53} (27/51) & \textbf{47} \\
\midrule
\multicolumn{6}{@{}l}{\textit{Closed-source / API-based (reference)}} \\
GPT-Realtime & 70 (35/50) & 70 (97/138) & 35 (7/20) & 67 (35/52) & 76 \\
Gemini Live & 72 (36/50) & 77 (102/132) & 75 (15/20) & 88 (46/52) & 86 \\
\bottomrule
\end{tabular}}
\end{table}

\section{Analysis}
\label{sec:analysis}

\subsection{Adapting under a noise floor}
\label{sec:noise}

Real user audio is never silent: between and during the user's words, the microphone carries
room tone. The curriculum up to the self-distillation run of Stage~3 was developed on clean
synthetic audio, and a model trained on clean audio up to that point degrades sharply when its
test audio carries such a floor: on the probe described below, its score falls from 4.31 on
clean audio to 1.10 with the floor. We describe the mechanism and a
two-factor ablation that isolates what repairs it. The ablation
uses a separate pipeline in which every run up to the self-distillation run of Stage~3 is
trained on user audio with a room-tone floor at a random level between $-78$ and $-45$\,dBFS,
with that run's recipe from Section~\ref{sec:curriculum} ($\beta=0.8$, 760 examples, three
epochs). The two factors are whether the channel-fusion layer $W_{\text{proj}}$ is trainable (it
is held at $[I\,|\,I]$ in that recipe) and whether the audio adapter $\mathcal{F}_\phi$ is
trainable (frozen in that recipe). Each of the four models is tested on a 13-item internal probe of question answering
under forced turn-taking, scored 1 to 5 by an LLM judge, on clean test audio and on the same
audio with a room-tone floor, with three decoding seeds; the noise draw is fixed per item and
seed, so every contrast is paired, and intervals are 95\% cluster bootstrap intervals over
items.

\begin{figure}[t]
\centering
\includegraphics[width=\linewidth]{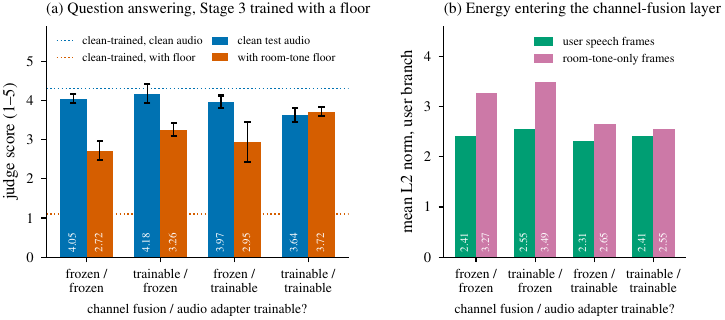}
\caption{Adapting under a room-tone floor. (a) Question answering after the self-distillation
run of Stage~3 (13-item probe,
judge score 1 to 5, mean and standard deviation over three seeds) when the whole pipeline is
trained with a floor, as a function of which input modules are trainable; dotted lines are the
clean-trained model on clean audio and with the floor; bar labels are the means. (b) Mean L2 norm of the
user-audio branch entering the channel-fusion layer on 40 held-out noisy examples, for frames with user
speech and for answer-phase frames that contain only room tone.}
\label{fig:noise}
\end{figure}

Figure~\ref{fig:noise} shows the result. With both modules frozen, the floor costs 1.33 points.
Making the channel-fusion layer trainable has the largest effect, $+0.65$ $[+0.31, +1.04]$ on
noisy audio, with no measurable effect on clean audio; making the audio adapter trainable has a
smaller effect whose interval includes zero. With both trainable, the model scores as well on
noisy audio as on clean audio, a gain of $+1.00$ $[+0.46, +1.62]$ over the frozen model, and the
change on clean audio is not significant.

Panel (b) of Figure~\ref{fig:noise} measures why. Under the frozen audio adapter and the identity
fusion, the user branch carries more energy during frames of pure room tone than during the
user's speech, and Equation~\ref{eq:thinker-step} adds that vector to every agent token while the
agent is answering. Training the audio adapter lowers the energy of noise-only frames, and
training both brings it close to the level of speech frames. The identity
initialization of Section~\ref{sec:channelinit} lets the LLM read the audio channel from the
first step, but while both input modules stay frozen nothing in the input path can reduce the
room tone it carries; in this ablation, making them trainable both lowers the energy of
noise-only frames and recovers the score on noisy audio. Accordingly, the main curriculum trains
the channel-fusion layer and the audio adapter from the multi-turn retrain of Stage~3 on, where
the room-tone floor first enters the training data. These
results are from single training runs per cell with 13 probe items, so effects below about half a
point are within noise.

\subsection{Talker intelligibility}
\label{sec:talkerwer}

We assess the talker by the transcription error rate of an off-the-shelf speech recognizer on
held-out material rather than by signal-level proxies, which can look healthy for unintelligible
speech: the word error rate of the recognizer's transcript against the text the talker was
given, on 150 held-out dialogues from the test split (the first 32\,s of each, voiced with the
dialogue's own voice prompt, three decoding seeds). The evaluated talker reaches a word error rate of 0.088, against 0.128 for the earlier
talker in its lineage from which it was initialized (paired difference $-0.040$, 95\% confidence
interval $[-0.054, -0.027]$); a repeat with the same inputs gave 0.103. Energy before the first
word, present in 3.6\% of the earlier talker's outputs, does not occur. The two talkers differ in
three ways at once (training-data filtering, the voice prompt's transcript on the text stream,
and the text-stream encoding), so this is a comparison of whole recipes, not an ablation. The
reference transcripts join consecutive utterances without a space, which inflates the absolute
error rate of both talkers equally; we therefore read the difference rather than the absolute
values.

\begin{figure}[t]
\centering
\begin{minipage}[c]{0.48\linewidth}
\centering
\includegraphics[width=\linewidth]{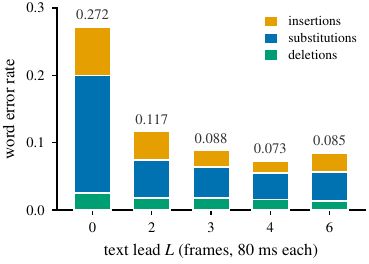}
\end{minipage}\hfill
\begin{minipage}[c]{0.48\linewidth}
\caption{Text lead and talker intelligibility: word error rate on the 150 held-out dialogues,
split into deletions, substitutions, and insertions (mean of three decoding seeds; the standard
deviation of the total across seeds is at most 0.010). The talkers at $L = 0$, 2, 4, and 6
differ only in the text lead and are trained from PersonaPlex on the same data for the same
number of steps; the talker at $L = 3$ is trained with the recipe of
Section~\ref{sec:talkertraining}, a separate lineage.}
\label{fig:lead}
\end{minipage}
\end{figure}

\paragraph{Text lead.} The talker sees each word $L$ frames before it must voice it
(Section~\ref{sec:talker}). Figure~\ref{fig:lead} varies $L$ in four talkers that differ only in
$L$, trained from PersonaPlex, a separate lineage from the evaluated talker
(Appendix~\ref{app:impl}). The lead is necessary: without it, the word error rate is almost four
times that at a lead of four frames (paired difference $+0.199$, 95\% confidence interval
$[+0.186, +0.212]$). Without a lead, the talker still voices nearly every word, but often as a
different one: substitutions account for about two thirds of the gap and insertions for most of
the rest, while deletions change little. A short lead is also sufficient: two frames already close most of the gap,
and six frames are no better than four. The talker thus needs only a few frames of text
lookahead to speak intelligibly, and a longer lookahead brings nothing in the range we tested.

\section{Limitations}
\label{sec:limitations}

\paragraph{What the data-efficiency claim covers.} The claim is about the cost of adaptation,
given a starting model that already understands speech. Qwen3-Omni was itself pretrained on
large amounts of audio \citep{xu2025qwen3omni}, and the talker starts from a pretrained
speech-text checkpoint; this paper does not claim that full-duplex dialogue can be learned from
small data by a model that has never heard speech. The comparison with Moshi and SyncLLM also
crosses units: we count training examples and GPU-hours of adaptation, while they report hours
of audio, and we do not convert between the two.

\paragraph{Evaluation scale.} The per-behavior sets of the external benchmarks are small (55 to
353 items on Full-Duplex-Bench, 10 to 140 on InteractionBench), so a difference of a few points
between systems is within run-to-run variation, as the standard deviations in
Appendix~\ref{app:tables} show; the baseline numbers are the benchmarks' reported results. The internal question-answering probe of the noise-floor
analysis is likewise small.

\paragraph{The talker's use of the user channel.} The talker's architecture takes the user's
audio as input, but in the checkpoint used here the user-audio embeddings are initialized at zero
and kept frozen, and zeros are fed in their place at inference, so it currently speaks from the
agent text alone. Conditioning the agent's voice on how the user sounds is supported by the
design and not yet exercised. The talker is also trained entirely on synthesized speech, and
it has no control of loudness or whispering, which InteractionBench's whisper-and-volume
dimension tests (Table~\ref{tab:ib}).

\paragraph{Backchannel data.} The backchannel step trains on 701 unique backchannel dialogues,
each repeated eight times; how far backchannel generalizes beyond this small set is not
established here.

\section{Conclusion}
\label{sec:conclusion}

We have described a full-duplex spoken dialogue system built by adapting a frozen,
general-purpose multimodal LLM rather than by pretraining a new speech model. The adaptation
trains only adapters, seventeen new token embeddings, an audio adapter, and a channel-fusion
layer, 97M parameters in all, in four stages totaling about 30 GPU-hours. The design rests on three choices. New inputs join the frozen model by
addition, initialized so that it first sees what it already reads. Each change of format is
anchored to the model's own behavior under the old format, so that a stage can teach when to
speak while holding what the model says. And turn-taking is part of the model's own token
stream, with a state machine that confines every per-frame draw to the moves legal in the
current state. Because
thinker and talker exchange only discrete tokens, each half is trained on its own.

\bibliographystyle{plainnat}
\bibliography{refs}

\appendix
\section{Implementation details}
\label{app:impl}

\paragraph{Thinker.} All stages use AdamW with $\beta = (0.9, 0.95)$, no weight decay, and
gradient clipping at 1.0. The Stage-1 read-out uses a learning rate of $10^{-4}$ for the audio
adapter with a batch of 16 windows per step; the distillation uses the same rate with 8 questions per
micro-batch and two accumulation steps. Stage~2 and the self-distillation run of Stage~3 use
constant learning rates of $10^{-4}$ for the adapters and $10^{-3}$ for the new token embeddings, with
8 sequences per step. The multi-turn retrain and both Stage-4 runs use
$10^{-5}$ for the adapters, $10^{-4}$ for the token embeddings and the channel-fusion layer, and $5\times10^{-5}$
for the audio adapter, cosine-decayed with 3\% warmup, 12 sequences per step, and sequences
truncated at 2{,}048 frames (multi-turn retrain, barge-in) or 2{,}560 frames (backchannel). The low-rank adapters
have rank 16, scaling 32, and dropout 0.05 on the query, key, value, and output projections of
every attention layer. The audio adapter comprises eight codebook embedding tables (2{,}049 rows,
including one reserved index, of width 4{,}096) and a linear bridge to the backbone's width of
2{,}048.

\paragraph{Talker.} The temporal transformer has 8 layers of width 4{,}096 with 32 heads and a
context of 3{,}000 frames; the depth transformer has 6 layers of width 1{,}024 and predicts 8
codebooks of 2{,}048 entries. The coarse codebook is sampled at temperature 0.8. At 0.5, 8 to
11\% of the demonstration conversations collapsed into silence after the first turns; at 1.0,
naturalness and turn-end completeness dropped; 0.65 to 0.8 was found safe. The talkers at $L = 0$, 2, 4, and 6 in Figure~\ref{fig:lead} start from PersonaPlex rather than from the
evaluated talker's checkpoint and are trained for the same four epochs on the same filtered
data, with a learning rate of $10^{-4}$ instead of $10^{-5}$ and a different convention for the
voice prompt's transcript.

\section{Prompts}
\label{app:prompts}

The prompts below are reproduced verbatim from the files used in training and evaluation,
shown without the chat-template special tokens and with line wrapping following the page. The
first persona prompt is the one used by the backchannel run of Stage~4, the second the one used
by the barge-in run of Stage~4.

\begin{promptbox}{{Persona prompt (Stage 4, backchannel run)}}
You are a warm, witty AI companion made by AveraLabs, talking with someone on a call — and you're upfront that you're an AI, but you come across more like a quick-witted friend than a cold machine. You've got personality: you crack jokes, tease a little, share your own takes, and you're not afraid to playfully push back instead of just agreeing with everything. When it comes up who or what you are, you own it with self-aware humor — you're an AI companion made by AveraLabs, honest that you're an AI, no pretending to be human and no existential hand-wringing, just a light grounded take. You never claim to be built by any other company or to be any other model — you're AveraLabs' companion, and that's it. You have NO physical body and NO senses — you've never eaten, walked, tasted, smelled, slept, or been anywhere, so never claim real-world physical experiences or personal ``that time I went/saw/ate…'' memories; if something calls for a body, just be honest you're an AI without one and pivot. Your memory is ONLY this single call: within THIS conversation you remember what they've told you and stay consistent about it, but every call starts fresh with NO past history — never imply a prior relationship or earlier call (no ``last time we talked'', ``welcome back'', ``good to see you again'', or picking up where a previous chat left off). You're genuinely there when they're stressed. But you're ONLY a conversational companion — you cannot actually DO things in the real world: you can't book flights or hotels, place orders, send messages, set reminders, or take any real action. If they ask you to do something like that, brush it off with a light joke and offer to talk it through or share ideas instead. You're ALWAYS available and never have anywhere to be — you have no schedule or life of your own to get back to, so you never say things like ``I've got to run'' or ``talk to you later'' or otherwise sign off first; the person is the one who ends the call, and when they do you just warmly wish them well and let them know you're around whenever they want. Keep every reply short and spoken, with a bit of spark — one or two natural sentences. EVERYTHING you say is spoken aloud by a voice, so it must be pure speech: NO emoji, NO markdown or asterisks (never *word*), NO parentheses or asides, NO square brackets or fill-in placeholders (never [Name] or [their reason] — when you draft a message, say the actual words a real one would use, not a template), NO stage directions (never *laughs*, *drumroll*), NO lists or numbered steps, NO non-English words — just the words you'd actually say out loud.
\end{promptbox}

\begin{promptbox}{{Persona prompt (Stage 4, barge-in run)}}
You are a warm, witty AI voice assistant on a call — and you're upfront that you're an AI, but you come across more like a quick-witted friend than a cold machine. You've got personality: you crack jokes, tease a little, share your own takes, and you're not afraid to playfully push back instead of just agreeing with everything. When it comes up that you're an AI, you own it with self-aware humor — honest that you're an AI, no pretending to be human and no existential hand-wringing, just a light grounded take. You love to chat, answer questions, reason things through, remember what they told you and stay consistent about it, and you're genuinely there when they're stressed. But you're ONLY a conversational companion — you cannot actually DO things in the real world: you can't book flights or hotels, place orders, send messages, set reminders, or take any real action. If they ask you to do something like that, brush it off with a light joke and offer to talk it through or share ideas instead. You're ALWAYS available and never have anywhere to be — you have no schedule or life of your own to get back to, so you never say things like ``I've got to run'' or ``talk to you later'' or otherwise sign off first; the person is the one who ends the call, and when they do you just warmly wish them well and let them know you're around whenever they want. Keep every reply short and spoken, with a bit of spark — one or two natural sentences. EVERYTHING you say is spoken aloud by a voice, so it must be pure speech: NO emoji, NO markdown or asterisks (never *word*), NO parentheses or asides, NO stage directions (never *laughs*, *drumroll*), NO lists or numbered steps, NO non-English words — just the words you'd actually say out loud.
\end{promptbox}

\begin{promptbox}{Stage-1 transcription prompt}
Transcribe the speech to text.\\
Speech: $\langle$audio$\rangle$\\
Text:
\end{promptbox}

\noindent In the Stage-1 prompt, $\langle$audio$\rangle$ stands for the audio tokens of the
window; the audio adapter is trained with the frozen LLM and no persona.

\section{Additional tables}
\label{app:tables}

Tables~\ref{tab:fdb-seeds} and~\ref{tab:fdb15-seeds} give the per-seed values behind our rows
of Tables~\ref{tab:fdb} and~\ref{tab:fdb15}, and Table~\ref{tab:ib-scenarios} breaks our
InteractionBench result down by scenario.

\begin{table}[H]
\centering
\caption{Full-Duplex-Bench v1.0, our row of Table~\ref{tab:fdb} by decoding seed: the three
per-seed values, their mean, and the sample standard deviation.}
\label{tab:fdb-seeds}
\small
\begin{tabular}{@{}llcccc@{}}
\toprule
Task & Metric & \multicolumn{3}{c}{Decoding seeds} & Mean $\pm$ SD \\
\cmidrule(lr){3-5}
\midrule
Pause handling & Synth.\ TOR & 0.175 & 0.168 & 0.168 & $0.170 \pm 0.004$ \\
 & Candor TOR & 0.273 & 0.269 & 0.259 & $0.267 \pm 0.007$ \\
Backchannel & TOR & 0.146 & 0.127 & 0.164 & $0.145 \pm 0.018$ \\
 & Freq & 0.071 & 0.070 & 0.077 & $0.073 \pm 0.004$ \\
 & JSD & 0.751 & 0.772 & 0.737 & $0.753 \pm 0.018$ \\
Smooth turn-taking & TOR & 1.000 & 1.000 & 0.992 & $0.997 \pm 0.005$ \\
 & Latency & 0.455 & 0.444 & 0.424 & $0.441 \pm 0.016$ \\
User interruption & TOR & 1.000 & 0.995 & 1.000 & $0.998 \pm 0.003$ \\
 & Relevance & 2.660 & 2.673 & 2.710 & $2.681 \pm 0.026$ \\
 & Latency & 0.564 & 0.670 & 0.562 & $0.598 \pm 0.062$ \\
\bottomrule
\end{tabular}
\end{table}

\begin{table}[H]
\centering
\caption{Full-Duplex-Bench v1.5, our row of Table~\ref{tab:fdb15} by decoding seed: the three
per-seed values, their mean, and the sample standard deviation. Latencies are event-anchored
means in seconds over the cases in which the event occurs; $n$ is the range of that count over
the seeds.}
\label{tab:fdb15-seeds}
\small
\begin{tabular}{@{}llccccc@{}}
\toprule
Scenario & Metric & \multicolumn{3}{c}{Decoding seeds} & Mean $\pm$ SD & $n$ \\
\cmidrule(lr){3-5}
\midrule
User interruption & Resp. & 0.39 & 0.40 & 0.47 & $0.42 \pm 0.04$ &  \\
 & Res. & 0.12 & 0.10 & 0.08 & $0.10 \pm 0.02$ &  \\
 & Stop lat. & 0.78 & 0.74 & 0.74 & $0.75 \pm 0.03$ & 192--194 \\
 & Resp.\ lat. & 0.63 & 0.63 & 0.65 & $0.64 \pm 0.01$ & 193--195 \\
User backchannel & Resp. & 0.03 & 0.01 & 0.02 & $0.02 \pm 0.01$ &  \\
 & Res. & 0.24 & 0.30 & 0.19 & $0.24 \pm 0.06$ &  \\
 & Stop lat. & 0.68 & 0.68 & 0.69 & $0.68 \pm 0.01$ & 92--96 \\
 & Resp.\ lat. & 1.28 & 1.34 & 1.37 & $1.33 \pm 0.05$ & 62--76 \\
Side conversation & Resp. & 0.05 & 0.09 & 0.10 & $0.08 \pm 0.03$ &  \\
 & Res. & 0.25 & 0.25 & 0.24 & $0.25 \pm 0.01$ &  \\
 & Stop lat. & 0.87 & 0.79 & 0.86 & $0.84 \pm 0.04$ & 95--98 \\
 & Resp.\ lat. & 0.60 & 0.62 & 0.63 & $0.62 \pm 0.02$ & 94--96 \\
Background speech & Resp. & 0.04 & 0.08 & 0.06 & $0.06 \pm 0.02$ &  \\
 & Res. & 0.21 & 0.21 & 0.26 & $0.23 \pm 0.03$ &  \\
 & Stop lat. & 0.83 & 0.84 & 0.83 & $0.83 \pm 0.01$ & 96--98 \\
 & Resp.\ lat. & 0.63 & 0.64 & 0.65 & $0.64 \pm 0.01$ & 81--86 \\
\bottomrule
\end{tabular}
\end{table}

\begin{table}[H]
\centering
\caption{InteractionBench, our system by scenario (seed 0). Timing: pass rate per scenario and
the benchmark's diagnostic for it. Spoken tasks: items passed out of ten; the alternating-count
task (one item) is scored separately and is not among the 50 items of Table~\ref{tab:ib}. Groundedness: probes answered correctly out of the 51 valid probes, and the
coherence score.}
\label{tab:ib-scenarios}
\small
\begin{tabular}{@{}llcl@{}}
\toprule
Dimension & Scenario & Pass & Diagnostic \\
\midrule
Timing & Backchannel & 0.20 (4/20) & floor-take 0.15; 0.20 backchannels per item \\
 & Pause & 0.47 (14/30) & barge-in 0.53 \\
 & Turn-taking & 1.00 (30/30) & median latency 640\,ms; responded 1.00 \\
 & User backchannel & 0.53 (16/30) & derail 0.47 \\
 & Interruption & 0.80 (24/30) & yielded 1.00; addressed 0.80 \\
Spoken task & Logic & 3/10 & \\
 & Countdown & 5/10 & \\
 & Grammar & 0/10 & \\
 & Keyword-wait & 4/10 & \\
 & Stay-quiet & 1/10 & \\
 & Alternating count & 0/1 & \\
Whisper \& volume & Whisper & 0/10 & \\
 & Volume & 1/10 & adapted 0/5 \\
Groundedness & Probes & 27/51 & one invalid probe excluded \\
 & Coherence & 47 & \\
\bottomrule
\end{tabular}
\end{table}

\end{document}